%% file: cvpr2019.tex
\documentclass[10pt,twocolumn,letterpaper]{article}

\usepackage{cvpr}
\usepackage{times}
\usepackage{epsfig}
\usepackage{graphicx}
\usepackage{amsmath}
\usepackage{amssymb}
\usepackage{times}  
\usepackage{helvet}  
\usepackage{courier}  
\usepackage{url}  
\usepackage{graphicx}  
\usepackage{multirow}
\usepackage{tabularx} 
\usepackage{caption}
\usepackage{xspace}
\usepackage{booktabs}
\usepackage{subfigure}
\usepackage{amsmath,amssymb} 

\input{tex/macros}

\usepackage[pagebackref=true,breaklinks=true,letterpaper=true,colorlinks,bookmarks=false]{hyperref}

\cvprfinalcopy 

\def\cvprPaperID{3750} 

\ifcvprfinal\fi
\begin{document}

\title{Revisiting Perspective Information for Efficient Crowd Counting}

\author{Miaojing Shi$^1$,~~Zhaohui Yang$^2$,~~Chao Xu$^2$,~~Qijun Chen$^3$ \\
	$^1$Univ Rennes, Inria, CNRS, IRISA \\
	$^2$Key Laboratory of Machine Perception, Cooperative Medianet Innovation Center, Peking University\\
	$^3$Department of Control Science and Engineering, Tongji University\\
}

\maketitle
\input{tex/sec-abstract.tex}

	\input{tex/sec-introduction.tex}

	\input{tex/sec-relatedworknew.tex}

	\input{tex/sec-method.tex}

	\input{tex/sec-experiment.tex}

	\input{tex/sec-conclusion.tex}

{\small
\bibliographystyle{ieee}
\bibliography{bibtex/eccv2018}
}

\end{document}

%% file: tex/macros.tex
\def \ie {{i.e.}\xspace}
\def \eg {{e.g.}\xspace}
\def \etc {{etc.}\xspace}
\def \etal {{et al.}\xspace}

\newcommand{\para}[1]{\noindent \textbf{#1}}



%% file: tex/sec-abstract.tex
\begin{abstract}
Crowd counting is the task of estimating people numbers in crowd images.
Modern crowd counting methods employ deep neural networks to estimate crowd counts via crowd density regressions.
A major challenge of this task lies in the perspective distortion, which results in drastic person scale change in an image. Density regression on the small person area is in general very hard.
In this work, we propose a perspective-aware convolutional neural network (PACNN) for efficient crowd counting,
which integrates the perspective information into density regression to provide additional knowledge of the person scale change in an image.
Ground truth perspective maps are firstly generated for training; PACNN is then specifically designed to
predict multi-scale perspective maps, and encode them as perspective-aware weighting layers in the network to adaptively combine the outputs of multi-scale density maps.
The weights are learned at every pixel of the maps such that the final density combination is robust to the perspective distortion. 
We conduct extensive experiments on the ShanghaiTech, WorldExpo'10, UCF\_CC\_50, and UCSD datasets,
and demonstrate the effectiveness and efficiency of PACNN
over the 
state-of-the-art. 
\end{abstract}

%% file: tex/sec-introduction.tex
\section{Introduction}\label{Sec:Intro}
\begin{figure}[t]
	\centering
	\includegraphics[width=1\columnwidth]{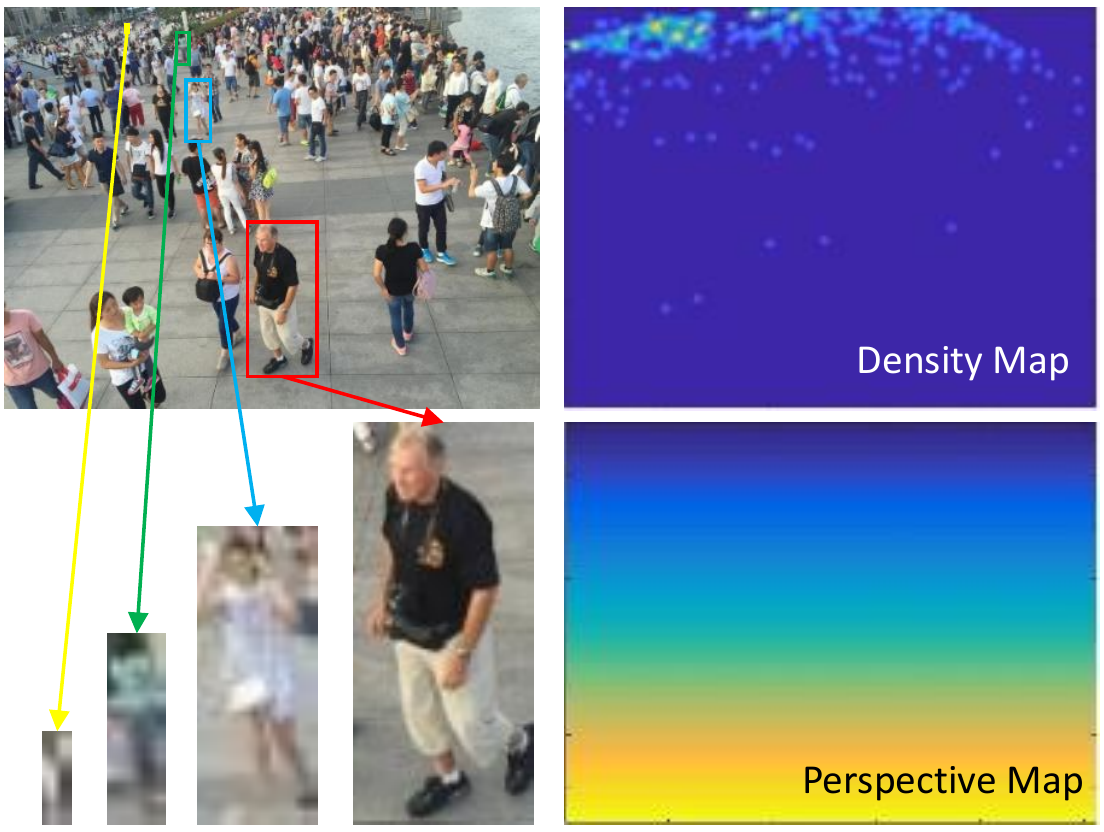}
	\caption{\small
		The density map shows the locally smoothed crowd count at every location in the image. The perspective map reflects the perspective distortion at every location in the image, \eg  how many pixels correspond to a human height of one meter at each location~\cite{zhang2015cvpr}. Person scale changes drastically due to the perspective distortion. Density regression on the small person area is in general very hard. We integrate the perspective map into density regression to provide additional information about the general person scale change from near to far in the image.
	}
	\label{Fig:intro}
		\vspace{-0.4cm}
\end{figure}

The rapid growth of the world's population has led to fast urbanization and resulted in more frequent crowd gatherings, \eg sport events, music festivals, political rallies. Accurate and fast crowd counting thereby
becomes essential to handle large crowds for public safety.
Traditional crowd counting methods estimate crowd counts via the detection of each individual pedestrian~\cite{wu2005iccv,viola2003ijcv,brostow2006cvpr,rabaud2006cvpr, liu2019cvpr}. Recent methods conduct crowd counting via the regression of density maps~\cite{chan2008cvpr,chen2012bmvc,ryan2009dicta,idrees2013cvpr}: the problem of crowd counting is casted as estimating a continuous density function whose integral over an image gives the count of persons within that image~\cite{chen2012bmvc,kong2005bmvc,lempitsky2010nips,onoro2016eccv,zhang2015cvpr,zhang2016cvpr,sam2017arxiv} (see Fig.~\ref{Fig:intro}: Density Map).
Handcrafted features were firstly employed in the density regression~\cite{chen2012bmvc,kong2005bmvc,lempitsky2010nips} and soon outperformed by deep representations~\cite{onoro2016eccv,zhang2015cvpr,zhang2016cvpr}.

A major challenge of this task lies in the drastic perspective distortions in crowd images (see Fig.~\ref{Fig:intro}).
The perspective problem is related to camera calibration which estimates a camera's 6 degrees-of freedom (DOF)~\cite{gao2003complete}.
Besides the camera DOFs, it is also defined in way to signify the person scale change from near to far in an image in crowd counting task
\cite{chan2008cvpr,chan2012tip,zhang2015cvpr,huang2018tip}. Perspective information has been widely used in traditional crowd counting methods to normalize features extracted at different locations of the image~\cite{chan2008cvpr,lempitsky2010nips,fiaschi2012icpr,loy2013msvac}. Despite the great benefits achieved by using image perspectives, there exists one clear disadvantage regarding its acquisition, which normally requires additional information/annotations of the camera parameters or scene geometries. The situation becomes serious when the community starts to employ deep learning to solve the problem in various scenarios~\cite{zhang2016cvpr,idrees2013cvpr}, where the perspective information is usually unavailable or not easy to acquire. 
While some works propose certain simple ways to label the perspective maps~\cite{chan2008cvpr,zhang2015cvpr}, most researchers in recent trends work towards a perspective-free setting~\cite{onoro2016eccv} where they exploit the multi-scale architecture of convolutional neural networks (CNNs) to regress the density maps at different resolutions~\cite{zhang2016cvpr,walach2016eccv,onoro2016eccv,sindagi2017iccv,sam2017arxiv,ranjan2018eccv,cao2018eccv}. To account for the varying person scale and crowd density, the patch-based estimation scheme~\cite{zhang2015cvpr,onoro2016eccv,sindagi2017iccv,sam2017arxiv,xiong2017iccv,liu2018cvpr,shen2018cvpr,cao2018eccv} is usually
adopted such that different patches are predicted (inferred) with different contexts/scales in the network. The improvements are significant but time costs are also expensive.


In this work, we revisit the perspective information for efficient crowd counting. We show that, with a little effort on the perspective acquisition, we are able to generate perspective maps for varying density crowds. We propose to integrate the perspective maps into crowd density regression to provide additional information about the person scale change in an image, which is particularly helpful on the density regression of small person area. The integration directly operates on the pixel-level, such that the proposed approach can be both efficient and accurate. To summarize, we propose a perspective-aware CNN (PACNN) for crowd counting. The contribution of our work concerns two aspects regarding the perspective generation and its integration with crowd density regression:

\textbf{(A)} The ground truth perspective maps are firstly generated for network training: sampled perspectives are computed at several person locations
based on their relations to person size;
a specific nonlinear function is proposed to fit the sampled values in each image based on the perspective geometry.
Having the ground truth, we train the network to directly predict perspective maps for new images.

\textbf{(B)} The perspective maps are explicitly integrated into the network to guide the multi-scale density combination:
three outputs are adaptively combined via two perspective-aware weighting layers in the network , where the weights in each layer are learned through a nonlinear transform of the predicted perspective map at the corresponding resolution. The final output is robust to the perspective distortion; we thereby infer the crowd density over the entire image.

We conduct extensive experiments on several standard benchmarks \ie ShanghaiTech~\cite{zhang2016cvpr}, WorldExpo'10~\cite{zhang2015cvpr}, UCF\_FF\_50~\cite{idrees2013cvpr} and UCSD~\cite{chan2008cvpr}, to show the superiority of our PACNN over the state-of-the-art.

%% file: tex/sec-relatedworknew.tex
\section{Related work}
We categorize the literature in crowd counting into traditional and modern methods. Modern methods refer to those employ CNNs while traditional methods do not.  

\subsection{Traditional methods}
\para{Detection-based methods.} These methods consider a crowd as a group of detected individual pedestrians~\cite{liu2019cvpr, wang2011cvpr,wu2005iccv,stewart2016cvpr,viola2003ijcv,brostow2006cvpr,rabaud2006cvpr}.
They can be performed either in a monolithic manner or part-based.
Monolithic approach typically refers to pedestrian detection that employs hand-crafted features like Haar~\cite{viola2001cvpr} and HOG~\cite{dalal2005cvpr} to train an SVM or AdaBoost detector~\cite{stewart2016cvpr,viola2003ijcv,brostow2006cvpr,rabaud2006cvpr}. These approaches often perform poorly in the dense crowds where pedestrians are heavily occluded or overlapped. Part-based detection is therefore adopted in many works~\cite{lin2001tsmc,wang2011cvpr,wu2005iccv,idrees2015pami} to count pedestrian from parts in images.
Despite the improvements achieved, the detection-based crowd counting overall suffers severely in dense crowds with complex backgrounds.

\para{Regression-based methods.} These methods basically have two steps: first, extracting effective features from crowd images; second, utilizing various regression functions to estimate crowd counts. Regression features include edge features~\cite{chan2008cvpr,chen2012bmvc,ryan2009dicta,regazzoni1996sp,chan2012tip}, texture features~\cite{chen2012bmvc,idrees2013cvpr,marana1998sibgrapi,chan2012tip} \etc Regression methods include linear~\cite{regazzoni1996sp,paragios2001cvpr}, ridge~\cite{chen2012bmvc} and Gaussian~\cite{chan2008cvpr,chan2012tip} functions.
Earlier works ignore the spatial information by simply regressing a scalar value (crowd count), later works instead learn a mapping from local features to a density map~\cite{chen2012bmvc,kong2005bmvc,lempitsky2010nips}. Spatial locations of persons are encoded into the density map; the crowd count is obtained by integrating over the density map.

\emph{Perspective information} was widely used in traditional crowd counting methods, which provides additional information regarding the person scale change along with the perspective geometry. It is usually utilized to normalize the regression features or detection results~\cite{chan2008cvpr,lempitsky2010nips,loy2013msvac,idrees2015pami}.

\subsection{Modern methods}\label{Sec:CNNMethods}
Due to the use of strong CNN features, recent works on crowd counting have shown remarkable progress~\cite{zhang2015cvpr,boominathan2016mm,zhang2016cvpr,onoro2016eccv,zhao2016eccv,sindagi2017avss,sindagi2017iccv,sam2017arxiv,xiong2017iccv,zhang2018wacv,liu2018cvpr,liu2018bcvpr,li2018cvpr,shi2018cvpr,shen2018cvpr,cao2018eccv,ranjan2018eccv}.
In order to deal with the varying head size in one image, the multi-column~\cite{zhang2016cvpr,onoro2016eccv,sam2017arxiv,boominathan2016mm} or multi-scale~\cite{zhang2018wacv,cao2018eccv,shen2018cvpr,ranjan2018eccv} network architecture is often utilized for crowd density regression. Many works also adopt a patch-based scheme to divide each image into local patches corresponding to different crowd densities and scales~\cite{onoro2016eccv,sam2017arxiv,shen2018cvpr,cao2018eccv}.
For example, \cite{onoro2016eccv} uses a pyramid of image patches extracted at multiple scales and feeds them into different CNN columns; while Sam~\etal~\cite{sam2017arxiv} introduce a switch classifier to relay the crowd patches from images to their best CNN columns with most suitable scales.
~Sindagi~\etal \cite{sindagi2017iccv} design a system called contextual pyramid CNN.
It consists of both a local and global context estimator to perform patch-based density estimation.
Shen~\etal~\cite{shen2018cvpr} introduce an adversarial loss to generate density map for sharper and higher resolution and design a novel scale-consistency regularizer which enforces that the sum of the crowd counts
from local patches is coherent with the
overall count of their region union.
Cao~\etal~\cite{cao2018eccv} propose a novel encoder-decoder network and local pattern consistency loss in crowd counting.
A patch-based test scheme is also applied to reduce the impact of statistic shift problem.


\emph{Perspective information} was also used in modern methods but often in an implicit way, \eg to normalize the scales of pedestrians in the generalization of ground truth density~\cite{zhang2015cvpr,zhang2016cvpr} or body part~\cite{huang2018tip} maps. We instead predict the perspective maps directly in the network and use them to adaptively combine the multi-scale density outputs. 
There are also other works trying to learn or leverage different cues to address the perspective distortion in images~\cite{idrees2015pami,arteta2016eccv}.
For instance,~\cite{idrees2015pami} uses locally-consistent scale prior maps to detect and count humans in dense crowds;
while~\cite{arteta2016eccv} employs a depth map to predict the size of objects in the wild and count them. 



%% file: tex/sec-method.tex
\section{Perspective-aware CNN}\label{Sec:Method}

\begin{figure}[t]
	\centering
	\includegraphics[width=0.75\columnwidth]{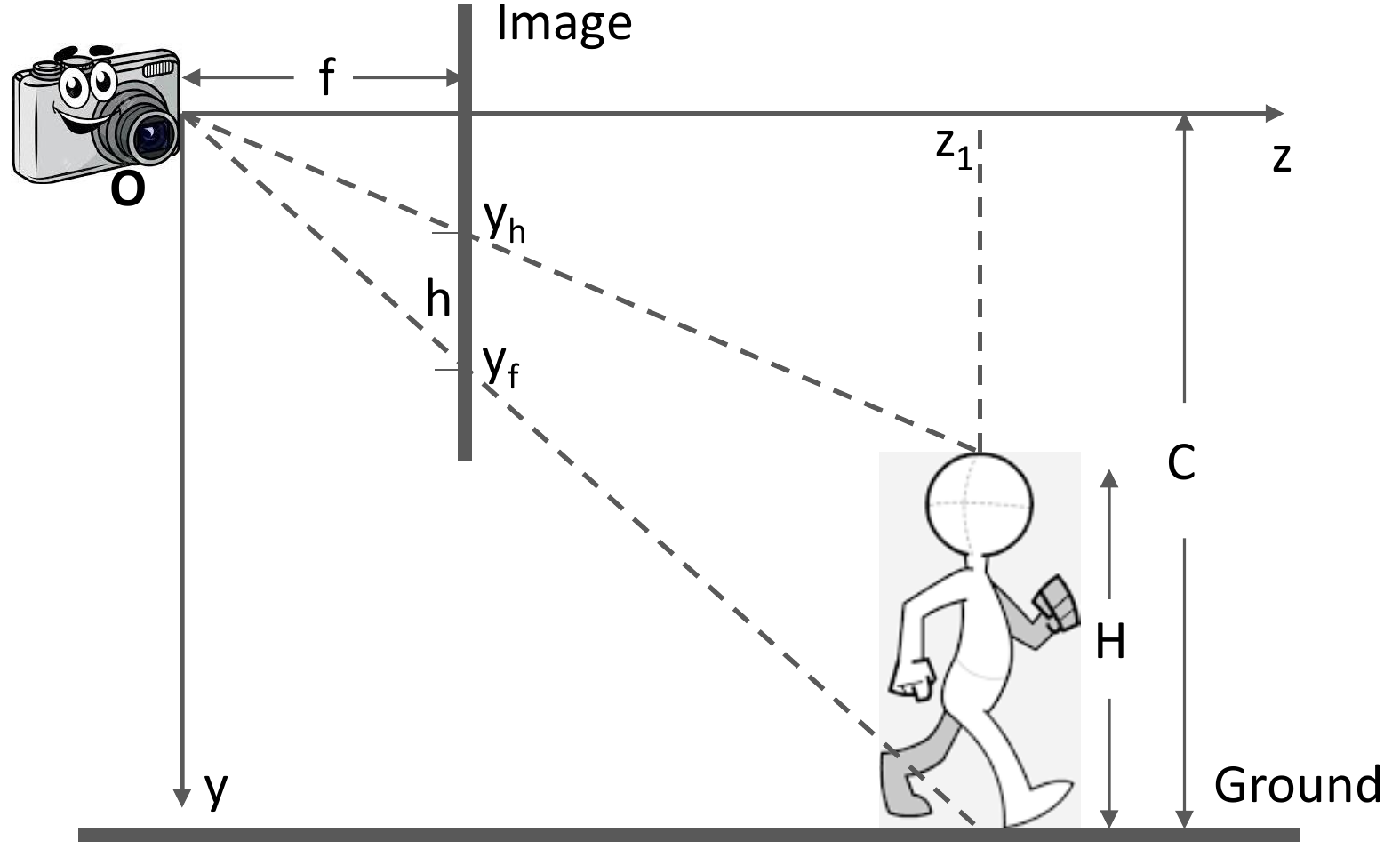}
	\caption{\small The perspective geometry of a pinhole camera  seen from the $x$-axis. The Cartesian coordinate system starts from origin $\textbf O$, with $y$-axis representing the vertical direction while $z$-axis the optical axis (depth). 
	A person with true height $H$ is walking on the ground, and he is shot by a camera located at $\textbf O$ where the camera aperture is. The person's head top and feet bottom are mapped on the image plane at $y_h$ and $y_f$, respectively. The distance from the camera aperture to the image plane is $f$, which is also known as the focal length. The camera height from the ground is $C$.
	}
	\label{Fig:perspective}
	\vspace{-0.3cm}
\end{figure}

In this section we first generate ground truth density maps and perspective maps; then introduce the network architecture;
finally present the network training protocol.

\subsection{Ground truth (GT) generation}\label{Sec:GTmaps}
\para{GT density map generation.} 
The GT density map $D^g$ can be generated by convolving Gaussian kernel $G_{\sigma}$ with head center annotation $z_j$, as in~\cite{zhang2016cvpr,sam2017arxiv,sindagi2017iccv}:
\begin{equation}\label{Eq:Gaussian}
D^g = \sum\limits_{j = 1}^{Y^g} G_{\sigma}(z-z_j),
\end{equation}
where $Y^g$ denotes the total number of persons in an image; $\sigma$ is obtained following~\cite{zhang2016cvpr}. The integral of $D^g$ is equivalent to $Y^g$ (see Fig.~\ref{Fig:intro}).

\medskip

\para{GT perspective map generation.}
\textit{Perspective maps} were widely used in~\cite{chan2008cvpr,lempitsky2010nips,fiaschi2012icpr,loy2013msvac,zhang2015cvpr,huang2018tip}.
The GT perspective value at every pixel of the map $P^{g} = \{p^{g}_j\}$ is defined as the number of pixels representing one meter at that location in the real scene~\cite{zhang2015cvpr}. The observed object size in the image is thus related to the perspective value. Below we first review the conventional approach to compute the perspective maps in crowded scenes of pedestrians.

\begin{figure}[t]
	\centering
	\includegraphics[width=1\columnwidth]{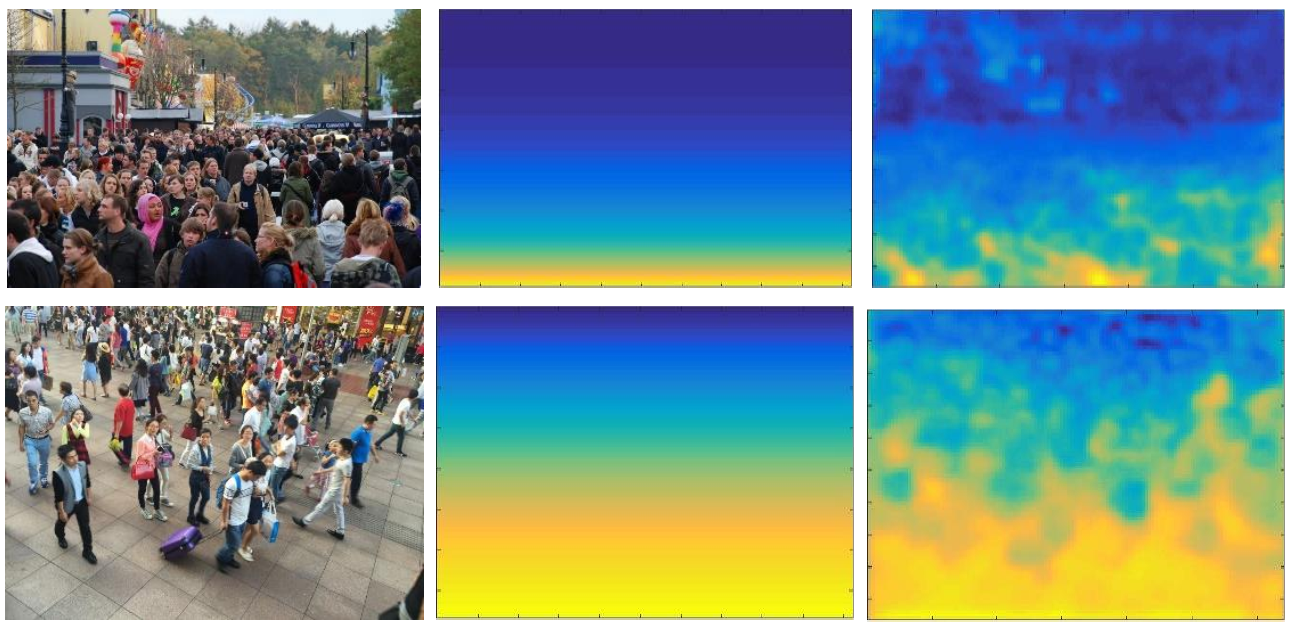}
	\caption{\small
		Perspective samples from SHA and SHB~\cite{zhang2016cvpr}.
In each row, the left column is the original image, middle column is the GT perspective map using (\ref{Eq:Perspective5}) while the right column is the estimated perspective map by PACNN.
		Blue in the heatmaps indicates small perspective values while yellow indicates large values.
		%
		%
	}
	\label{Fig:PMAP}
		\vspace{-0.4cm}
\end{figure}

\emph{Preliminary. } Fig.~\ref{Fig:perspective} visualizes the perspective geometry of a pinhole camera. Referring to the figure caption, we can solve the similar triangles,
 \begin{equation}\label{Eq:Perspective}
\begin{aligned}
y_h &= \frac{f(C-H)}{z_1},\\
y_f &= \frac{fC}{z_1},
\end{aligned}
\end{equation}
where $y_h$ and $y_f$ are the observed positions of person head and feet on the image plane, respectively. The observed person height $h$ is thus given by,
\begin{equation}\label{Eq:Perspective2}
h = y_f - y_h = \frac{fH}{z_1}
\end{equation}
dividing the two sides of (\ref{Eq:Perspective2}) by $y_h$ will give us
\begin{equation}\label{Eq:Perspective3}
h = \frac{H}{C-H} y_h.
\end{equation}
The perspective value $p^g$ is therefore defined as:
\begin{equation}\label{Eq:Perspective4}
p^g = \frac{h}{H} = \frac{1}{C-H} y_h.
\end{equation}
To generate the perspective map for a crowd image, authors in~\cite{zhang2015cvpr} approximate $H$ to be the mean height of adults (1.75m) for every pedestrian. Since $C$ is fixed for each image, $p^g$ becomes a linear function of $y_h$ and remains the same in each row of $y_h$. To estimate $C$, they manually labeled the heights $h_j$ of several pedestrians at different positions in each image, such that the perspective value $p^g_j$ at the sampled position $j$ is given by $p^{g}_j = \frac{{{h_j}}}{{1.75}}$.
They employ a linear regression method afterwards to fit Eqn. (\ref{Eq:Perspective4}) and generate the entire GT perspective map.


\medskip

The perspective maps for datasets WorldExpo'10~\cite{zhang2015cvpr} and UCSD~\cite{chan2008cvpr} were generated via the above process. However, for datasets having dense crowds like ShanghaiTech PartA (SHA)~\cite{zhang2016cvpr} and UCF\_CC\_50~\cite{idrees2013cvpr}, it can not directly apply as the pedestrian bodies are usually not visible in dense crowds.
We notice that, similar to the observed pedestrian height, the head size also changes with the perspective distortion.
We therefore interpret the sampled perspective value $p^{g}_j$ by the observed head size, which can be computed following~\cite{zhang2016cvpr} as the average distance from certain head at $j$ to its K-nearest neighbors (K-NN). 


The next step is to generate the perspective map based on the sampled values. The conventional linear regression approach~\cite{chan2008cvpr,zhang2015cvpr} relies on several assumptions, \eg the camera is not in-plane rotated; the ground in the captured scene is flat; the pedestrian height difference is neglected; and most importantly, the sampled perspective values are accurate enough. The first three assumptions are valid for many images in standard crowd counting benchmarks, but there exist special cases such that the camera is slightly rotated; people sit in different tiers of a stadium; and the pedestrian height (head size) varies significantly within a local area. As for the last, using the K-NN distance to approximate the pedestrian head size is surely not perfect; noise exists even in dense crowds as the person distance highly depends on the local crowd density at each position.

Considering the above facts, now we introduce a novel nonlinear way to fit the perspective values, aiming to produce an accurate perspective map that clearly underlines the general head size change from near to far in the image. First, we compute the mean perspective value at each sampled row $y_h$ so as to reduce the outlier influence due to any abrupt density or head size change. We employ a $\tanh$ function to fit these mean values over their row indices $y_h$:
\begin{equation}\label{Eq:Perspective5}
p^g = a \cdot \tanh(b \cdot(y_h+c)),
\end{equation}
where $a$, $b$ and $c$ are three parameters to fit in each image. This function produces a perspective map with values decreased from bottom to top and identical in the same row, indicating the vertical person scale change in the image. 

The local distance scale has been utilized before to help normalize the detection of traditional method~\cite{idrees2015pami}; while in modern CNN-based methods, it is often utilized implicitly in the ground truth density generation~\cite{zhang2016cvpr}. Unlike in~\cite{idrees2015pami,zhang2016cvpr}, the perspective is more than the local distance scale:
we mine the reliable perspective information from sampled local scales and fit a nonlinear function over them,
which indeed provides additional information about person scale change at every pixel due to the perspective distortion. Moreover, we explicitly encode the perspective map into CNN to guide the density regression at different locations of the image (as below described).
The proposed perspective maps are not yet perfect but demonstrated to be helpful (see Sec.~\ref{Sec:Experiment}). On the other hand, if we simply keep the K-NN distance as the final value in the map, we barely get no significant benefit in our experiment.

We generate the GT perspective maps for datasets UCF\_CC\_50 and ShanghaiTech SHA using our proposed way. While for SHB,
the pedestrian bodies are normally visible and the sampled perspective values can be simply obtained by labeling several (less than 10) pedestrian heights; unlike the conventional way, the nonlinear fitting procedure (\ref{Eq:Perspective5}) is still applied. We illustrate some examples in Fig.~\ref{Fig:PMAP} for both SHA and SHB. Notice we also evaluate the linear regression for GT perspective maps in crowd counting, which performs lower than our non-linear way.

\begin{figure*}[t]
	\centering
	\includegraphics[width=0.9\textwidth]{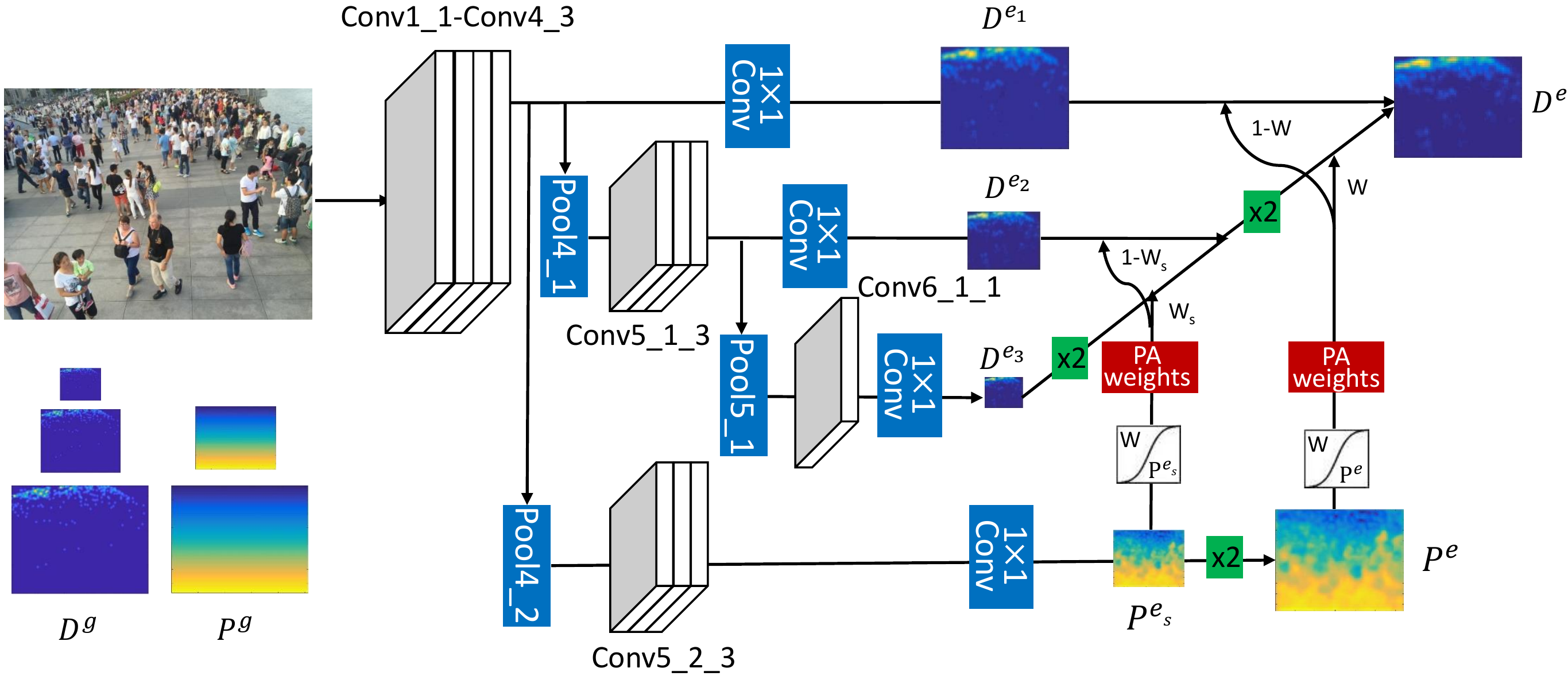}
	\caption{
		\small The structure of the proposed perspective-aware convolutional neural network (PACNN).
		$D$ and $P$ denote the density and perspective map,
		while $e$ and $g$ stand for estimation and ground truth; green box ``$x2$" denotes the deconvolutional layer for upsampling.
		The backbone is adopted from the VGG net.
		We regress three density maps $D^{e_1}$, $D^{e_2}$ and $D^{e_3}$ from Conv4\_3, Conv5\_1\_3 and Conv6\_1\_1, respectively;
       two perspective maps $P^{e_s}$ and $P^e$ are produced after Conv5\_2\_3.
		We adaptively combine the multi-scale density outputs via two perspective-aware (PA) weighting layers,
		where the PA weights are learned via the nonlinear transform of $P^{e_s}$ and $P^e$.
		We optimize the network over the loss with respect to the ground truth of $D^{g}$ and $P^g$ in different resolutions.
		The final density output is $D^e$.
	}
	\label{Fig:network}
		\vspace{-0.4cm}
\end{figure*}

\subsection{Network architecture}\label{Sec:Architecture}
We show the network architecture in Fig.~\ref{Fig:network}: the backbone is adopted from the VGG net~\cite{simonyan2015iclr};
out of Conv4\_3, we branch off several data streams to perform the density and perspective regressions, which are described next.

\para{Density map regression.} 
We regress three density maps from the outputs of Conv4\_3, Conv5\_1\_3 and Conv6\_1\_1 simultaneously. The filters from deeper layers have bigger receptive fields than those from the shallower layers. Normally, a combination of the three density maps is supposed to adapt to varying person size in an image.

We denote by $D^{e_1} = \{d^{e_1}_j\}$, $D^{e_2}= \{d^{e_2}_j\}$ and $D^{e_3} = \{d^{e_3}_j\}$ the three density maps from Conv4\_3, Conv5\_1\_3, and Conv6\_1\_1, respectively; $j$ signifies the $j$-th pixel in the map; they are regressed using 1$\times$1 Conv with 1 output.
Because of pooling, $D^{e_1}$, $D^{e_2}$, and $D^{e_3}$ have different size: $D^{e_1}$ is of $1/8$ resolution of the input, while $D^{e_2}$ and $D^{e_3}$ are of $1/16$ and $1/32$ resolutions, respectively. We downsample the ground truth density map to each corresponding resolution to learn the multi-scale density maps. To combine them, a straightforward way would be averaging their outputs: $D^{e_3}$ is firstly upsampled via a deconvoltuional layer to the same size with  $D^{e_2}$; we denote it by  $\mathrm{Up}(D^{e_3})$, $\mathrm{Up}(\cdot)$ is the  deconvolutional upsampling; we average $\mathrm{Up}(D^{e_3})$ and $D^{e_2}$ as $(D^{e_2} + \mathrm{Up}(D^{e_3}))/2$; the averaged output is upsampled again and further combined with $D^{e_1}$ to produce the final output $D^{e}$:
 \begin{equation}\label{Eq:DensityAvg}
    D^{e} = \frac{D^{e_1} + \mathrm{Up}(\frac{D^{e_2} + \mathrm{Up}(D^{e_3})}{2})}{2}
 \end{equation}
$D^{e}$ is of $1/8$ resolution of the input, and we need to downsample the corresponding ground truth as well.
This combination is a simple, below we introduce our perspective-aware weighting scheme. 

\para{Perspective map regression.} 
Perspective maps are firstly regressed in the network. The regression is branched off from Pool4\_2 with three more convolutional layers Conv5\_2\_1 to Conv5\_2\_3. We use $P^{e_s} = \{p^{e_s}_j\}$ to denote the regressed perspective map after Conv5\_2\_3.
It is with 1/16 resolution of the input, we further upsample it to 1/8 resolution of the input to obtain the final perspective map $P^{e} = \{p^{e}_j\}$.  We prepare two perspective maps $P^{e_s}$ and $P^{e}$ to separately combine the output of $D^{e_2}$ and $\mathrm{Up}(D^{e_3})$, as well as  $\mathrm{Up}(D^{e_2} + \mathrm{Up}(D^{e_3}))/2$ and $D^{e_1}$ at different resolutions.
Ground truth perspective map is downsampled accordingly to match the estimation size.
We present some estimated perspective maps $P^e$ and their corresponding ground truths $P^g$ in Fig.~\ref{Fig:PMAP}. 

\para{Perspective-aware weighting.} Due to different receptive field size, $D^{e_1}$ is normally good at estimating small heads, $D^{e_2}$ medium heads, while $D^{e_3}$ big heads.
We know that the person size in general decreases with an decrease of the perspective value. To make use of the estimated perspective maps $P^{e_s}$ and $P^{e}$, we add two perspective-aware (PA) weighting layers in the network (see Fig.~\ref{Fig:network}) to specifically adapt the combination of $D^{e_1}$, $D^{e_2}$ and $D^{e_3}$ at two levels. The two PA weighting layers work in a similar way to give a density map higher weights on the smaller head area if it is good at detecting smaller heads, and vice versa. We start by formulating the combination between $D^{e_2}$ and $\mathrm{Up}(D^{e_3})$:
 \begin{equation}\label{Eq:DensityWeight}
    D^{e_s} = W^s \odot D^{e_2} + (1 - W^s) \odot \mathrm{Up}(D^{e_3}),
 \end{equation}
where $\odot$ denotes the element-wise (Hadamard) product and $D^{e_s}$ the combined output. $W^s  = \{w^s_j\}$ is the output of the perspective-aware weighting layer; it is obtained by applying a nonlinear transform $w^s_j = f(p^{e_s}_j)$ to the perspective values $p^{e_s}_j$ (nonlinear transform works better than linear transform in our work).
This function needs to be differentiable and produce a positive mapping from $p^{e_s}_j$ to $w^s_j$. We choose the sigmoid function:
   \begin{equation}\label{Eq:Weight}
    w^s_j = f({p^{e_s}_j}) = \frac{1}{{1 + {\exp({-\alpha^s* ({p^{e_s}_j} - \beta^s ))}}}},
 \end{equation}
where $\alpha^s$ and $\beta^s$ are the two parameters that can be learned via back propagation. $w^s_j \in (0,1)$, it varies at every pixel of the density map. The backwards function of the PA weighting layer computes partial derivative of the loss function $L$ with respect to $\alpha^s$ and $\beta^s$. We will discuss the loss function later. Here we write out the chain rule:
\begin{equation}\label{Eqn: diffalpha}
\begin{aligned}
&\frac{{\partial {L}}}{{\partial \alpha^s }}  =  \frac{{\partial L}}{{\partial {D^{e_s}}}}\frac{{\partial {D^{e_s}}}}{{\partial W^s}}\frac{{\partial W^s}}{{\partial \alpha^s }}\\ 
 = & \sum\nolimits_j \frac{{\partial L}}{{\partial {d_j^{e_s}}}} {(d_j^{{e_2}} - \mathrm{Up}(d_j^{{e_3}}))(p_j^{e_s} - \beta^s )f(p_j^{e_s})(1 - f(p_j^{e_s}))};
\end{aligned}
\end{equation}
Similarly, we have
\begin{equation}\label{Eqn: diffbeta}
\begin{aligned}
\frac{{\partial {L}}}{{\partial \beta^s }} = \sum\nolimits_j \frac{{\partial L}}{{\partial {d_j^{e_s}}}} (d_j^{{e_2}} - \mathrm{Up}(d_j^{{e_3}}))( - \alpha^s )f(p_j^{e_s})(1 - f(p_j^{e_s})).
\end{aligned}
\end{equation}

The output $D^{e_s}$ can be further upsampled and combined with $D^{e_1}$ using another PA weighting layer:
 \begin{equation}\label{Eq:DensityLargeScale}
D^{e} = W \odot D^{e_1} + (1 - W) \odot \mathrm{Up}(D^{e_s}),
\end{equation}
where $W = \{w_j\}$ is transformed from $P^e$ in a similar way to $W^s$:
   \begin{equation}\label{Eq:WeightLS}
w_j = f({p^{e}_j}) = \frac{1}{{1 + {\exp({-\alpha* ({p^{e}_j} - \beta ))}}}},
\end{equation}
$\alpha$ and $\beta$ are two parameters similar to $\alpha^s$ and $\beta^s$ in (\ref{Eq:Weight}); one can follow (\ref{Eqn: diffalpha},\ref{Eqn: diffbeta}) to write out their backpropagations. Compared to the average operation in (\ref{Eq:DensityAvg}), which gives the same weights in the combination,
the proposed perspective-aware weighting scheme (\ref{Eq:DensityWeight}) 
gives different weights on $D^{e_1}$, $D^{e_2}$ and $D^{e_3}$ at different positions of the image, such that the final output is robust to the perspective distortion.

\subsection{Loss function and network training}\label{Sec:Loss}
We regress both the perspective and density maps in a multi-task network. In each specific task $\mathrm O$, a typical loss function is the mean squared error (MSE) loss $L^{\mathrm {MSE}}$, which sums up the pixel-wise Euclidean distance between the estimated map and ground truth map. The MSE loss does not consider the local correlation in the map, likewise in~\cite{cao2018eccv}, we adopt the DSSIM loss to measure the local pattern consistency between the estimated map and ground truth map. The DSSIM loss $L^{\mathrm {DSSIM}}$ is derived from the structural similarity (SSIM)~\cite{wang2004tip}. The whole loss for task $\mathrm O$ is thereby,
 \begin{equation}\label{Eq: Loss}
 \begin{split}
& {L}_{\mathrm O}(\Theta )  =  L^{\mathrm {MSE}} + \lambda L^{\mathrm {DSSIM}} \\
&= \frac{1}{2N}\sum\limits_{i = 1}^N \|E({X_i};\Theta ) - G_i\|_2^2 \\
& + \lambda   \frac{1}{N}\sum\limits_{i = 1}^N (1- \frac{1}{M}\sum_j\mathrm{SSIM}_i(j))\\
& \mathrm{SSIM}_i  =\frac{(2\mu_{E_i} \mu_{G_i} + C_1)}{\mu^2_{E_i} + \mu^2_{G_i} + C_1} \cdot \frac{(2\sigma_{{E_i} {G_i}} + C_2)}{\sigma^2_{E_i} + \sigma^2_{G_i} + C_2}
\end{split}
\end{equation}
where $\Theta$ is a set of learnable parameters in the proposed network; $X_i$ is the input image, $N$ is the number of training images and $M$ is the number of pixels in the maps; $\lambda$ is the weight to balance $L^{\mathrm {MSE}}$ and $L^{\mathrm {DSSIM}}$. We denote by $E$ and $G$ the respective estimated map and ground truth map for task $\mathrm O$. Means ($\mu_{E_i}$, $\mu_{G_i}$) and standard deviations ($\sigma_{E_i}$, $\sigma_{G_i}$, $\sigma_{E_iG_i}$) in SSIM$_i$ are computed with a Gaussian filter with standard deviation 1 within a $5\times5$ region at each position $j$. We omit the dependence of means and standard deviations on pixel $j$ in the equation.

\begin{figure}[t]
	\centering
	\includegraphics[width=1\columnwidth]{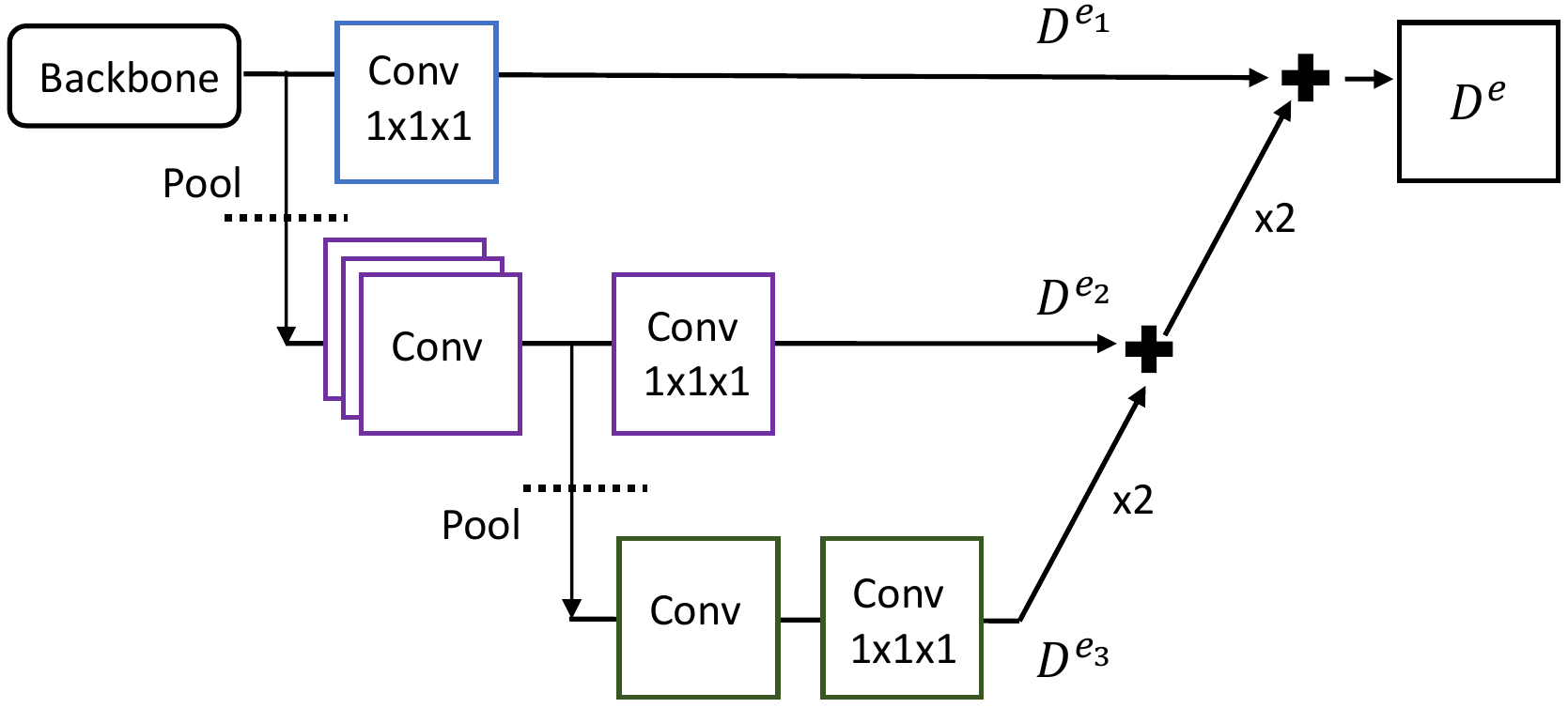}
	\caption{Network architecture without using perspective (denoted as PACNN w/o P). Referring to (\ref{Eq:DensityAvg}), multi-scale density outputs are adapted to the same resolution and averaged to produce the final prediction.
	}
	\label{Fig:concat}
	\vspace{-0.4cm}
\end{figure}

For the perspective regression task $\mathrm P$, we obtain its loss $L_{\mathrm P}$ from (\ref{Eq: Loss}) by substituting $P^{e}$ and $P^{g}$ into $E$ and $G$, respectively; while for the density regression task $\mathrm D$, we obtain its loss $L_{\mathrm D}$ by replacing $E$ and $G$ with $D^{e}$ and $D^{g}$ correspondingly. We offer our overall loss function as
\begin{equation}\label{Eq: finalLoss}
L = L_{\mathrm P} + L_{\mathrm D} + \kappa L_{\mathrm {P^s}} +  \lambda_1 L_{\mathrm {D^1}} + \lambda_2 L_{\mathrm{D^2}} +  \lambda_3 L_{\mathrm{D^3}}.
\end{equation}
As mentioned in Sec.~\ref{Sec:Architecture}, $L_{P^s}$ is a subloss for $P^{e_s}$ while $L_{D^1}$, $L_{D^2}$ and $L_{D^3}$ are the three sub-losses for $D^{e_1}$, $D^{e_2}$ and $D^{e_3}$. We empirically give small loss weights for these sublosses. We notice that the ground truth perspective and density maps are pre-processed to have the same scale in practice.  
The training is optimized with Stochastic Gradient Descent (SGD) in two phases. \textbf{Phase 1}:
we optimize the density regression using the architecture in Fig~\ref{Fig:concat}; \textbf{Phase 2}: we finetune the model by adding the perspective-aware weighting layers to jointly optimize the perspective and density regressions.

%% file: tex/sec-experiment.tex
\section{Experiments}\label{Sec:Experiment}

\subsection{Datasets}\label{Sec:Dataset}
\para{ShanghaiTech~\cite{zhang2016cvpr}.} It consists of 1,198 annotated images with a total of 330,165 people with head center annotations. This dataset is split into two parts SHA and SHB. The crowd images are sparser in SHB compared to SHA: the average crowd counts are 123.6 and 501.4, respectively. Following~\cite{zhang2016cvpr}, we use 300 images for training and 182 images for testing in SHA; 400 images for training and 316 images for testing in SHB.

\para{WorldExpo'10~\cite{zhang2015cvpr}.} It includes 3,980 frames, which are taken from the Shanghai 2010 WorldExpo. 3,380 frames are used as training while the rest are taken as test. The test set includes five different scenes and 120 frames in each one. Regions of interest (ROI) are provided in each scene so that crowd counting is only conducted in the ROI in each frame. The crowds in this dataset are relatively sparse with an average pedestrian number of 50.2 per image.

\para{UCF\_CC\_50~\cite{idrees2013cvpr}.} It has 50 images with 63,974 head annotations in total. The head counts range between 94 and 4,543 per image. The small dataset size and large count variance make it a very challenging dataset. Following~\cite{idrees2013cvpr}, we perform 5-fold cross validations to report the average test performance.

\para{UCSD~\cite{chan2008cvpr}.} This dataset contains 2000 frames chosen from one surveillance
camera in the UCSD campus. The frame size is
158 $\times$ 238 and it is recorded at 10 fps. There are only about
25 persons on average in each frame. It provides the ROI for each video frame. Following~\cite{chan2008cvpr}, we use frames
from 601 to 1400 as training data, and the remaining 1200
frames as test data.

\subsection{Implementation details and evaluation protocol}\label{Sec:ExperimentalDetails}
Ground truth annotations for each head center  are publicly available in the standard benchmarks. For WolrdExpo'10 and UCSD, the ground truth perspective maps are provided. For ShanghaiTech and UCF, ground truth perspective maps are generated as described in Sec.~\ref{Sec:GTmaps}\footnote{Ground truth perspective maps for ShanghaiTech can be downloaded from here:  \url{https://drive.google.com/open?id=117MLmXj24-vg4Fz0MZcm9jJISvZ46apK}}.
Given a training set, we augment it by randomly cropping 9 patches from each image. Each patch is $1/4$ size of the original image. All patches are used to train our PACNN. The backbone is adopted from VGG-16~\cite{simonyan2015iclr}, pretrained on ILSVRC classification data.
We set the batch size as 1, learning rate 1e-6 and momentum 0.9. We train 100 epochs in Phase 1 while 150 epochs in Phase 2 (Sec.~\ref{Sec:Loss}). Network inference is on the entire image.

We evaluate the performance via the mean absolute error (MAE) and mean squared error (MSE) as commonly used in previous works~\cite{zhang2015cvpr,zhang2016cvpr,sam2017arxiv,onoro2016eccv,wang2015mm}:
Small MAE and MSE values indicate good performance.

\subsection{Results on ShanghaiTech}\label{Sec:Ablation}
\para{Ablation study.} We conduct an ablation study to justify the utilization of multi-scale and perspective-aware weighting schemes in PACNN. Results are shown in Table~\ref{Tab:Shanghaitech}.


Referring to Sec.~\ref{Sec:Architecture}, $D^{e_1}$, $D^{e_1}$ and $D^{e_3}$  should fire more on small, medium and big heads, respectively. Having a look at Table~\ref{Tab:Shanghaitech}, the MAE for $D^{e_1}$ $D^{e_2}$ and $D^{e_3}$ on SHA are 81.8, 86.3 and 93.1, respectively; on SHB they are 16.0, 14.5 and 18.2, respectively. Crowds in SHA are much denser than in SHB, persons are mostly very small in SHA and medium/medium-small in SHB.
It reflects in Table~\ref{Tab:Shanghaitech} that $D^{e_1}$ in general performs better on SHA while $D^{e_2}$ performs better on SHB. 

To justify the PA weighting scheme, we compare PACNN with the average weighting scheme (see Fig.~\ref{Fig:concat}) in Table~\ref{Tab:Shanghaitech}. Directly averaging over pixels of $D^{e_1}$ and upsampled $D^{e_2}$ and $D^{e_3}$ (PACNN w/o P) produces a marginal improvement of MAE and MSE on SHA and SHB. For instance, the MAE is decreased to 76.5 compared to 81.8 of $D^{e_1}$ on SHA; 12.9 compared to 14.5 of $D^{e_2}$ on SHB.
In contrast, using PA weights to adaptively combine $D^{e_1}$, $D^{e_1}$ and $D^{e_3}$ significantly decreases the MAE and MSE on SHA and SHB: they are 66.3 and 106.4 on SHA; 8.9 and 13.5 on SHB, respectively. 


\begin{table}[t]
	\setlength{\tabcolsep}{2.6pt}
	\centering
	\small
\begin{tabular}{|c|c|c|c|c|c|}
          \hline
  \multirow{2}{*}{ShanghaiTech}  & \multirow{2}{*}{Inference} & \multicolumn{2}{c|}{SHA}& \multicolumn{2}{c|}{SHB} \\
     \cline{3-6}
    & &  MAE& MSE & MAE & MSE \\
     \hline
     \hline
       $D^{e_1}$  &image  & 81.8 & 131.1 & 16.0 & 21.9 \\
     \hline
            $D^{e_2}$&   image & 86.3 & 138.6 & 14.5 & 18.7 \\
     \hline
          $D^{e_3}$ &  image & 93.1 & 156.4 & 18.2 & 25.1 \\
     \hline
       PACNN w/o P  &  image &{76.5} & {123.3} & {12.9} & {17.2} \\
     \hline
      PACNN   & image & {66.3} & {106.4} & {8.9} & {13.5} \\
     \hline
       PACNN + \cite{li2018cvpr}  & image & \textbf{62.4} & \textbf{102.0} & \textbf{7.6} & \textbf{11.8} \\
     \hline
     \hline
     Cao~\etal \cite{cao2018eccv}   & patch & \textbf{67.0} & 104.5 & \textbf{8.4} & \textbf{13.6} \\
     \hline
          Ranjan~\etal~\cite{ranjan2018eccv} & image$^*$ & 68.5 & 116.2 & 10.7 & 16.0 \\
          \hline
     	Li~\etal \cite{li2018cvpr} & image & {68.2} & 115.0 & {10.6} & {16.0} \\
     \hline
     Liu~\etal~\cite{liu2018bcvpr} &  - & 73.6 & {112.0} & 13.7 & 21.4\\
     \hline
     Shen~\etal~\cite{shen2018cvpr}  & patch & 75.7 & \textbf{102.7}& 17.2& 27.4 \\
    \hline
     Sindagi~\etal~\cite{sindagi2017iccv}  & patch & {73.6} & {106.4} & {20.1} & {30.1} \\
     \hline
\end{tabular}
	\caption{\small Ablation study of PACNN and its comparison with state-of-the-art on ShanghaiTech dataset. $D^{e_1}$, $D^{e_2}$ and $D^{e_3}$ denote the density map regressed from Conv4\_3, Conv5\_1\_3 and Conv6\_1\_1 in Fig.~\ref{Fig:network}, respectively. ``Inference" signifies whether it is patch-based or image-based. ``-" means it is not mentioned in the paper. ``image$^*$" denotes that a two-stage inference in~\cite{ranjan2018eccv}. PACNN w/o P denotes our network without using perspective maps (see Fig.~\ref{Fig:concat}). }
	
	\label{Tab:Shanghaitech}	
     \vspace{-0.5cm}
\end{table}

\begin{figure*}[t]
	\centering
	\includegraphics[width=0.8\textwidth]{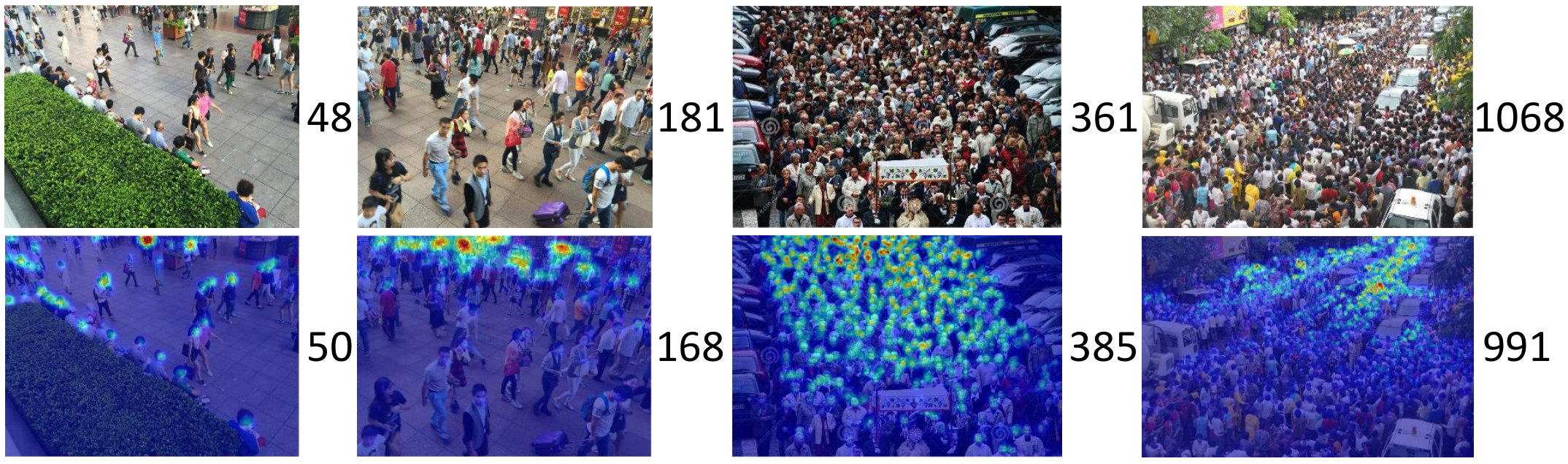}
	\caption{
	\small	Results on ShanghaiTech dataset. We present four test images and their estimated density maps below. The ground truth and estimated crowd counts are to the right of the real images and the corresponding density maps, respectively. }
	\label{Fig:shanghaitech}
\vspace{-0.2cm}
\end{figure*}

\para{Comparison to state-of-the-art.} We compare PACNN with the state-of-the-art~\cite{sindagi2017iccv,shen2018cvpr,liu2018bcvpr,li2018cvpr,ranjan2018eccv,cao2018eccv} in Table~\ref{Tab:Shanghaitech}.
PACNN produces the lowest MAE 66.3 on SHA and lowest MSE 13.5 on SHB, the second lowest MSE 106.4 on SHA and MAE 8.9 on SHB compared to the previous best results~\cite{cao2018eccv,shen2018cvpr}.  We notice that
many previous methods employ the patch-based inference~\cite{sindagi2017iccv,shen2018cvpr,cao2018eccv},
where model inference is usually conducted with a sliding window strategy.
We illustrate the inference type for each method in Table~\ref{Tab:Shanghaitech}.
Patch-based inference can be very time-consuming factoring the additional cost to crop and resize patches from images and merge their results.
On the other hand, PACNN employs an image-based inference and can be very fast; for instance,
in the same Caffe~\cite{jia2014mm} framework with an Nvidia GTX Titan X GPU, the inference time of our PACNN for an 1024*768 input is only 230ms while those with patch-based inference can be much (\eg 5x) slower in our experiment.
If we compare our result to previous best result with the image-based inference (\eg \cite{li2018cvpr}), ours is clearly better.
We can further combine our method with~\cite{li2018cvpr} by adopting its trained backbone,
we achieve the lowest MAE and MSE: 62.4 and 102.0 on SHA, 7.6 and 11.8 on SHB.
This demonstrates the robustness and efficiency of our method in a real application.
Fig.~\ref{Fig:shanghaitech} shows some examples. 

\subsection{Results on UCF\_CC\_50}
We compare our method with other state-of-the-art on UCF\_CC\_50~\cite{sindagi2017iccv,liu2018bcvpr,li2018cvpr,ranjan2018eccv,cao2018eccv} in Table~\ref{Tab:UCF}.
Our method PACNN achieves the MAE 267.9 and MSE 357.8; while the best MAE is 258.4 from~\cite{cao2018eccv} and  MSE 320.9 from ~\cite{sindagi2017iccv}.
We also present the result of PACNN + \cite{li2018cvpr}, which produces the lowest MAE and MSE: 241.7 and 320.7. We notice the backbone model of~\cite{li2018cvpr} that we use to combine with PACNN is trained by ourselves. Our reproduced model produces slightly lower MAE and MSE (262.5 and 392.7) than the results in~\cite{li2018cvpr}.

\begin{table}[t]
	\centering
	\small
\begin{tabular}{|c|c|c|}
          \hline
     UCF\_CC\_50 & MAE& MSE\\
     \hline
      Sindgai~\etal~\cite{sindagi2017iccv} & {295.8} & {320.9}\\
       Liu~\etal~\cite{liu2018bcvpr}  & 279.6 & 388.5 \\
       Li~\etal~\cite{li2018cvpr}  & 266.1 & 397.5 \\
      Ranjan~\etal~\cite{ranjan2018eccv}   & 260.9 & 365.5 \\
      Cao~\etal~\cite{cao2018eccv} & {258.4} & 344.9 \\
     PACNN & 267.9 & 357.8 \\
     PACNN + \cite{li2018cvpr} & \textbf{241.7} & \textbf{320.7} \\
     \hline
\end{tabular}
	\caption{\small Comparison of PACNN with other state-of-the-art on UCF\_CC\_50 dataset. }
	\label{Tab:UCF}
     \vspace{-0.1cm}
\end{table}

\begin{table}[t]
	\setlength{\tabcolsep}{3.6pt}
	\centering
	\small
	\begin{tabular}{|c|c|c|c|c|c|c|}
		\hline
		WorldExpo'10 & S1& S2 & S3 & S4 & S5 & Avg.\\
		\hline
		Sindagi~\etal~\cite{sindagi2017iccv}  &2.9  & 14.7 & 10.5 & 10.4 & 5.8 & 8.9 \\
			Xiong~\etal~\cite{xiong2017iccv}  &6.8  & 14.5 & 14.9 & 13.5 & 3.1 & 10.6 \\
				Li~\etal~\cite{li2018cvpr} & 2.9 & \textbf{11.5}& \textbf{8.6} & 16.6& 3.4  & 8.6\\
				Liu~\etal~\cite{liu2018bcvpr}  &\textbf{2.0}  & {13.1} & {8.9} & 17.4 & 4.8 & 9.2 \\
				Ranjan~\etal~\cite{ranjan2018eccv}& 17.0 & 12.3& 9.2& \textbf{8.1} & 4.7 & 10.3  \\
				Cao~\etal~\cite{cao2018eccv}& 2.6 & {13.2} & 9.0& 13.3 & \textbf{3.0} & 8.2\\
		PACNN & 2.3 & 12.5 & 9.1 & {11.2} & 3.8 & \textbf{7.8} \\
		\hline
	\end{tabular}
	\caption{\small Comparison of PACNN with other state-of-the-art on WorldExpo'10 dataset. MAE is reported for each test scene and averaged in the end. }
	\label{Tab:WorldExpo}
		\vspace{-0.3cm}
\end{table}

\subsection{Results on WorldExpo'10}
Referring to~\cite{zhang2015cvpr}, training and test are both conducted within the ROI provided for each scene of WorldExpo'10. MAE is reported for each test scene and averaged to evaluate the overall performance. We compare our PACNN with other state-of-the-art~\cite{sindagi2017iccv,xiong2017iccv,liu2018cvpr,li2018cvpr,ranjan2018eccv,cao2018eccv} in Table~\ref{Tab:WorldExpo}.
It can be seen that although PACNN does not outperform the state-of-the-art in each specific scene, it produces the lowest mean MAE 7.8 over the five scenes. Perspective information is in general helpful for crowd counting in various scenarios.

\begin{table}[t]
	\centering
	\small
	\begin{tabular}{|c|c|c|}
		\hline
		UCSD & MAE& MSE\\
		\hline
		Zhang~\etal~\cite{zhang2016cvpr} & 1.60 & 3.31 \\
		Onoro~\etal~\cite{onoro2016eccv} & {1.51} & - \\
		Sam~\etal~\cite{sam2017arxiv}  & 1.62 & 2.10 \\
		Huang~\etal~\cite{huang2018tip}  & 1.00 & 1.40 \\
		Li~\etal~\cite{li2018cvpr}   & 1.16& 1.47 \\
		Cao~\etal~\cite{cao2018eccv} & 1.02 & 1.29 \\
		PACNN & \textbf{0.89} & \textbf{1.18} \\
		\hline
	\end{tabular}
	\caption{\small Comparison of PACNN with other state-of-the-art on UCSD dataset. }
	\label{Tab:UCSD}
	\vspace{-0.4cm}
\end{table}

\subsection{Results on UCSD}
The crowds in this dataset is not evenly distributed and the person scale changes drastically due to the perspective distortion. Perspective maps were originally proposed in this dataset to weight each image location in the crowd segment according to its approximate size in the real scene. We evaluate our PACNN in Table~\ref{Tab:UCSD}: comparing to the state-of-the-art~\cite{zhang2016cvpr,onoro2016eccv,huang2018tip,sam2017arxiv,li2018cvpr,cao2018eccv}, PACNN significantly decreases the MAE and MSE to the lowest: 0.89 and 1.18, which demonstrates the effectiveness of our perspective-aware framework. Besides, the crowds in this dataset is in general sparser than in other datasets, which shows the generalizability of our method over varying crowd densities.

%% file: tex/sec-conclusion.tex
\section{Conclusion}
In this paper we propose a perspective-aware convolutional neural network to automatically estimate the crowd counts in images.
A novel way of generating GT perspective maps is introduced for PACNN training, such that at the test stage it predicts both the perspective maps and density maps. The perspective maps are encoded as two perspective-aware weighting layers to adaptively combine
the multi-scale density outputs. The combined density map is demonstrated to be robust to the perspective distortion in crowd images. Extensive experiments on standard crowd counting benchmarks show the efficiency and effectiveness of the proposed method over the state-of-the-art.

 
\medskip
\noindent \textbf{Acknowledgments.} This work was supported by NSFC 61828602 and 61733013. Zhaohui Yang and Chao Xu were supported by NSFC 61876007 and 61872012. We thank Dr. Yannis Avrithis for the discussion on perspective geometry and Dr. Holger Caesar for proofreading.

%% file: cvpr2019.bbl
\begin{thebibliography}{10}\itemsep=-1pt

\bibitem{arteta2016eccv}
C.~Arteta, V.~Lempitsky, and A.~Zisserman.
\newblock Counting in the wild.
\newblock In {\em ECCV}, 2016.

\bibitem{boominathan2016mm}
L.~Boominathan, S.~S. Kruthiventi, and R.~V. Babu.
\newblock Crowdnet: a deep convolutional network for dense crowd counting.
\newblock In {\em ACM MM}, 2016.

\bibitem{brostow2006cvpr}
G.~J. Brostow and R.~Cipolla.
\newblock Unsupervised bayesian detection of independent motion in crowds.
\newblock In {\em CVPR}, 2006.

\bibitem{cao2018eccv}
X.~Cao, Z.~Wang, Y.~Zhao, and F.~Su.
\newblock Scale aggregation network for accurate and efficient crowd counting.
\newblock In {\em ECCV}, 2018.

\bibitem{chan2008cvpr}
A.~B. Chan, Z.-S.~J. Liang, and N.~Vasconcelos.
\newblock Privacy preserving crowd monitoring: Counting people without people
  models or tracking.
\newblock In {\em CVPR}, 2008.

\bibitem{chan2012tip}
A.~B. Chan and N.~Vasconcelos.
\newblock Counting people with low-level features and bayesian regression.
\newblock {\em IEEE Transactions on Image Processing}, 21(4):2160--2177, 2012.

\bibitem{chen2012bmvc}
K.~Chen, C.~C. Loy, S.~Gong, and T.~Xiang.
\newblock Feature mining for localised crowd counting.
\newblock In {\em BMVC}, 2012.

\bibitem{dalal2005cvpr}
N.~Dalal and B.~Triggs.
\newblock Histograms of oriented gradients for human detection.
\newblock In {\em CVPR}, 2005.

\bibitem{fiaschi2012icpr}
L.~Fiaschi, U.~K{\"o}the, R.~Nair, and F.~A. Hamprecht.
\newblock Learning to count with regression forest and structured labels.
\newblock In {\em ICPR}, pages 2685--2688, 2012.

\bibitem{gao2003complete}
X.-S. Gao, X.-R. Hou, J.~Tang, and H.-F. Cheng.
\newblock Complete solution classification for the perspective-three-point
  problem.
\newblock {\em IEEE Transactions on Pattern Analysis and Machine Intelligence},
  25(8):930--943, 2003.

\bibitem{huang2018tip}
S.~Huang, X.~Li, Z.~Zhang, F.~Wu, S.~Gao, R.~Ji, and J.~Han.
\newblock Body structure aware deep crowd counting.
\newblock {\em IEEE Transactions on Image Processing}, 27(3):1049--1059, 2018.

\bibitem{idrees2013cvpr}
H.~Idrees, I.~Saleemi, C.~Seibert, and M.~Shah.
\newblock Multi-source multi-scale counting in extremely dense crowd images.
\newblock In {\em CVPR}, 2013.

\bibitem{idrees2015pami}
H.~Idrees, K.~Soomro, and M.~Shah.
\newblock Detecting humans in dense crowds using locally-consistent scale prior
  and global occlusion reasoning.
\newblock {\em TPAMI}, 37(10):1986--1998, 2015.

\bibitem{jia2014mm}
Y.~Jia, E.~Shelhamer, J.~Donahue, S.~Karayev, J.~Long, R.~Girshick,
  S.~Guadarrama, and T.~Darrell.
\newblock Caffe: Convolutional architecture for fast feature embedding.
\newblock In {\em ACM MM}, 2014.

\bibitem{kong2005bmvc}
D.~Kong, D.~Gray, and H.~Tao.
\newblock Counting pedestrians in crowds using viewpoint invariant training.
\newblock In {\em BMVC}, 2005.

\bibitem{lempitsky2010nips}
V.~Lempitsky and A.~Zisserman.
\newblock Learning to count objects in images.
\newblock In {\em NIPS}, 2010.

\bibitem{li2018cvpr}
Y.~Li, X.~Zhang, and D.~Chen.
\newblock Csrnet: Dilated convolutional neural networks for understanding the
  highly congested scenes.
\newblock In {\em CVPR}, 2018.

\bibitem{lin2001tsmc}
S.-F. Lin, J.-Y. Chen, and H.-X. Chao.
\newblock Estimation of number of people in crowded scenes using perspective
  transformation.
\newblock {\em TSMC-A}, 31(6):645--654, 2001.

\bibitem{liu2018cvpr}
J.~Liu, C.~Gao, D.~Meng, and A.~G.~Hauptmann.
\newblock Decidenet: Counting varying density crowds through attention guided
  detection and density estimation.
\newblock In {\em CVPR}, 2018.

\bibitem{liu2018bcvpr}
X.~Liu, J.~Weijer, and A.~D. Bagdanov.
\newblock Leveraging unlabeled data for crowd counting by learning to rank.
\newblock In {\em CVPR}, 2018.

\bibitem{liu2019cvpr}
Y.~Liu, M.~Shi, Q.~Zhao, and X.~Wang.
\newblock Point in, box out: Beyond counting persons in crowds.
\newblock In {\em CVPR}, 2019.

\bibitem{loy2013msvac}
C.~C. Loy, K.~Chen, S.~Gong, and T.~Xiang.
\newblock Crowd counting and profiling: Methodology and evaluation.
\newblock In {\em Modeling, Simulation and Visual Analysis of Crowds}, pages
  347--382. Springer, 2013.

\bibitem{zhang2018wacv}
Z.~Lu, M.~Shi, and Q.~Chen.
\newblock Crowd counting via scale-adaptive convolutional neural network.
\newblock In {\em WACV}, 2018.

\bibitem{marana1998sibgrapi}
A.~Marana, L.~d.~F. Costa, R.~Lotufo, and S.~Velastin.
\newblock On the efficacy of texture analysis for crowd monitoring.
\newblock In {\em SIBGRAPI}, 1998.

\bibitem{onoro2016eccv}
D.~Onoro-Rubio and R.~J. L{\'o}pez-Sastre.
\newblock Towards perspective-free object counting with deep learning.
\newblock In {\em ECCV}, 2016.

\bibitem{paragios2001cvpr}
N.~Paragios and V.~Ramesh.
\newblock A mrf-based approach for real-time subway monitoring.
\newblock In {\em CVPR}, 2001.

\bibitem{rabaud2006cvpr}
V.~Rabaud and S.~Belongie.
\newblock Counting crowded moving objects.
\newblock In {\em CVPR}, 2006.

\bibitem{ranjan2018eccv}
V.~Ranjan, H.~Le, and M.~Hoai.
\newblock Iterative crowd counting.
\newblock {\em ECCV}, 2018.

\bibitem{regazzoni1996sp}
C.~S. Regazzoni and A.~Tesei.
\newblock Distributed data fusion for real-time crowding estimation.
\newblock {\em Signal Processing}, 53(1):47--63, 1996.

\bibitem{ryan2009dicta}
D.~Ryan, S.~Denman, C.~Fookes, and S.~Sridharan.
\newblock Crowd counting using multiple local features.
\newblock In {\em DICTA}, 2009.

\bibitem{sam2017arxiv}
D.~B. Sam, S.~Surya, and R.~V. Babu.
\newblock Switching convolutional neural network for crowd counting.
\newblock In {\em CVPR}, 2017.

\bibitem{shen2018cvpr}
Z.~Shen, Y.~Xu, B.~Ni, M.~Wang, J.~Hu, and X.~Yang.
\newblock Crowd counting via adversarial cross-scale consistency pursuit.
\newblock In {\em CVPR}, 2018.

\bibitem{shi2018cvpr}
Z.~Shi, L.~Zhang, Y.~Liu, X.~Cao, Y.~Ye, M.-M. Cheng, and G.~Zheng.
\newblock Crowd counting with deep negative correlation learning.
\newblock In {\em CVPR}, 2018.

\bibitem{simonyan2015iclr}
K.~Simonyan and A.~Zisserman.
\newblock Very deep convolutional networks for large-scale image recognition.
\newblock In {\em ICLR}, 2015.

\bibitem{sindagi2017avss}
V.~A. Sindagi and V.~M. Patel.
\newblock Cnn-based cascaded multi-task learning of high-level prior and
  density estimation for crowd counting.
\newblock In {\em AVSS}, 2017.

\bibitem{sindagi2017iccv}
V.~A. Sindagi and V.~M. Patel.
\newblock Generating high-quality crowd density maps using contextual pyramid
  cnns.
\newblock In {\em ICCV}, 2017.

\bibitem{stewart2016cvpr}
R.~Stewart, M.~Andriluka, and A.~Y. Ng.
\newblock End-to-end people detection in crowded scenes.
\newblock In {\em CVPR}, 2016.

\bibitem{viola2001cvpr}
P.~Viola and M.~Jones.
\newblock Rapid object detection using a boosted cascade of simple features.
\newblock In {\em CVPR}, 2001.

\bibitem{viola2003ijcv}
P.~Viola, M.~J. Jones, and D.~Snow.
\newblock Detecting pedestrians using patterns of motion and appearance.
\newblock {\em IJCV}, 63(2):153--161, 2003.

\bibitem{walach2016eccv}
E.~Walach and L.~Wolf.
\newblock Learning to count with cnn boosting.
\newblock In {\em ECCV}, 2016.

\bibitem{wang2015mm}
C.~Wang, H.~Zhang, L.~Yang, S.~Liu, and X.~Cao.
\newblock Deep people counting in extremely dense crowds.
\newblock In {\em ACM MM}, 2015.

\bibitem{wang2011cvpr}
M.~Wang and X.~Wang.
\newblock Automatic adaptation of a generic pedestrian detector to a specific
  traffic scene.
\newblock In {\em CVPR}, 2011.

\bibitem{wang2004tip}
Z.~Wang, A.~C. Bovik, H.~R. Sheikh, and E.~P. Simoncelli.
\newblock Image quality assessment: from error visibility to structural
  similarity.
\newblock {\em IEEE transactions on image processing}, 13(4):600--612, 2004.

\bibitem{wu2005iccv}
B.~Wu and R.~Nevatia.
\newblock Detection of multiple, partially occluded humans in a single image by
  bayesian combination of edgelet part detectors.
\newblock In {\em ICCV}, 2005.

\bibitem{xiong2017iccv}
F.~Xiong, X.~Shi, and D.-Y. Yeung.
\newblock Spatiotemporal modeling for crowd counting in videos.
\newblock In {\em ICCV}, 2017.

\bibitem{zhang2015cvpr}
C.~Zhang, H.~Li, X.~Wang, and X.~Yang.
\newblock Cross-scene crowd counting via deep convolutional neural networks.
\newblock In {\em CVPR}, 2015.

\bibitem{zhang2016cvpr}
Y.~Zhang, D.~Zhou, S.~Chen, S.~Gao, and Y.~Ma.
\newblock Single-image crowd counting via multi-column convolutional neural
  network.
\newblock In {\em CVPR}, 2016.

\bibitem{zhao2016eccv}
Z.~Zhao, H.~Li, R.~Zhao, and X.~Wang.
\newblock Crossing-line crowd counting with two-phase deep neural networks.
\newblock In {\em ECCV}, 2016.

\end{thebibliography}
